# JANUS: A Dual-Constraint Generative Framework for Stealthy Node Injection Attacks

Jiahao Zhang[1], Xiaobing Pei[1], Zhaokun Zhong[1], Wenqiang Hao[1], Zhenghao Tang[1]

[1]School of Software Engineering, Huazhong University of Science and Technology
{jiahao_zhang, pei_xiaobing, zzk_hust, fivecoins, zh_tang42}@hust.edu.cn

## Abstract

Graph Neural Networks (GNNs) have demonstrated remarkable performance across various applications, yet they are vulnerable to sophisticated adversarial attacks, particularly node injection attacks. The success of such attacks heavily relies on their stealthiness, the ability to blend in with the original graph and evade detection. However, existing methods often achieve stealthiness by relying on indirect proxy metrics, lacking consideration for the fundamental characteristics of the injected content, or focusing only on imitating local structures, which leads to the problem of local myopia. To overcome these limitations, we propose a dual-constraint stealthy node injection framework, called **J**oint **A**lignment of **N**odal and **U**niversal **S**tructures (**JANUS**). At the local level, we introduce a local feature manifold alignment strategy to achieve geometric consistency in the feature space. At the global level, we incorporate structured latent variables and maximize the mutual information with the generated structures, ensuring the injected structures are consistent with the semantic patterns of the original graph. We model the injection attack as a sequential decision process, which is optimized by a reinforcement learning agent. Experiments on multiple standard datasets demonstrate that the JANUS framework significantly outperforms existing methods in terms of both attack effectiveness and stealthiness.

## Introduction

Graph Neural Networks (GNNs) have achieved great success in numerous fields such as node classification (Khoshraftar and An 2024; Mahmoud et al. 2024), graph classification (Liu, Chen, and Wen 2023; Khemani et al. 2024), recommendation systems (Gao et al. 2023; Sharma et al. 2024), and bioinformatics (Paul et al. 2024; Dong et al. 2023). However, with the widespread deployment of GNNs in security-sensitive domains like finance, social media, and critical infrastructure networks, their security issues have become increasingly prominent (Guan et al. 2024). A large number of studies have shown that GNNs are also vulnerable to adversarial attacks (Dai et al. 2024; Zhang et al. 2023). Attackers can significantly degrade the performance of GNN models, or even induce them to produce specific erroneous outputs, by modifying the topological structure of graphs (Hu et al. 2023; Wu et al. 2024) or injecting maliciously designed nodes into graphs (Sun et al. 2020; Zhu et al. 2024), which may lead to severe consequences in practical applications (Li et al. 2024; Zhao et al. 2023).

Among various attack paradigms, Graph Node Injection Attack has attracted much attention due to its unique practical feasibility (Zari et al. 2024). Injection attacks achieve the attack objective by injecting malicious nodes into graphs. This strategy usually has higher flexibility as it does not directly tamper with the protected original data.

For node injection attacks, stealthiness is a prerequisite for their successful deployment (Cai et al. 2024). No matter how theoretically destructive an easily detectable attack is, it will be ineffective in the real defense system (Chen et al. 2024). Although some progress has been made in the research on improving the stealthiness of injection attacks, there are generally two common limitations:

First, the modeling of local authenticity often relies on indirect and non-fundamental constraints. Many methods attempt to achieve stealthiness by maintaining certain specific attributes of the graph. HAO (Chen et al. 2022) aims to imitate the proxy metric of homophily, and GANI (Fang et al. 2024) generates features by sampling the degree of the original graph and based on the statistical data of the target category. These methods only consider the superficial silhouette of the real data distribution. And G-NIA (Tao et al. 2021), which uses the combination of original node features and scales node attributes to a preset range, is a heuristic post-processing. These methods all lack an end-to-end differentiable constraint that is directly aligned with the real data distribution.

Second, the consideration of global consistency is generally lacking, leading to the myopic problem of local optimization. Most existing methods focus on the local environment of the injected node. For example, CANA (Tao et al. 2023) improves local stealthiness by aligning the distribution of the ego-network of the injected node; while G²A2C (Ju et al. 2023) only constrains the injected features by limiting the feature budget. TDGIA (Zou et al. 2021) attacks the GNN model by using the topological defect edge selection strategy with the first-order neighborhood information of the graph. They all lack a constraint that ensures, at the architectural level, that multiple injection behaviors conform to the potential syntax rules of the graph at the global level, so that multiple locally normal injections may still accumulate into a perceptible global structural anomaly.

To overcome the aforementioned limitations, we propose a dual-constraint stealthy node injection framework, called **J**oint **A**lignment of **N**odal and **U**niversal **S**tructures (**JANUS**). We reframe the attack as a generative modeling problem of learning the distribution of original graph data. The core is an innovative dual stealthiness constraint mechanism, which systematically solves the dual challenges of local authenticity and global consistency:

1. Local node feature authenticity: At the node level, to address the limitations of proxy metrics, we propose a local feature manifold alignment strategy. By introducing the Optimal Transport (OT) (Peyré and Cuturi 2019) theory, we directly measure and align the original feature distribution, and minimize the transport cost between the empirical distribution of the features of the injected nodes and the empirical distribution of the features of the benign nodes sampled from the graph. Geometrically, this ensures that the injected features are a credible sample on the local feature manifold in a geometric sense, fundamentally improving their naturality in the feature space.

2. Global graph attribute consistency: To address the myopic problem of local optimization, we introduce a structure generation strategy based on controllable semantics. By extending the core idea of latent factor disentanglement in InfoGAN (Chen et al. 2016), we introduce a set of structured latent codes to control the generation process. By maximizing the mutual information between these latent codes and the generation results, we force the generator to learn the potential, high-level structural patterns and semantic rules in the original graph data.

The generation process of the entire attack is modeled as a sequential decision-making problem, which is optimized by a reinforcement learning agent under the guidance of the above-mentioned dual constraints. The main contributions of this paper can be summarized as follows:

- We propose a novel generative attack framework named JANUS. It systematically addresses the stealthiness challenge from two levels of local feature authenticity and global structure consistency for the first time.

- We design an end-to-end reinforcement learning attack framework. Under a unified optimization objective, it collaboratively realizes the learning of attack efficiency and dual stealthiness constraints.

- Through extensive experiments on multiple benchmark graph datasets, we verify that JANUS can achieve superior attack effects while significantly surpassing existing methods in multiple stealthiness indicators.

## Preliminaries

A graph is typically denoted as $G = (V, E, \mathbf{X})$, where $V$ is a set of nodes, $E$ is a set of edges, and $\mathbf{X} \in \mathbb{R}^{|V| \times d}$ is a node attribute matrix. The structure of a graph is usually represented by an adjacency matrix $\mathbf{A} \in \{0, 1\}^{|V| \times |V|}$.

### Graph Neural Networks

GNNs are models designed to process graph-structured data. Their core idea is to iteratively update node representations by aggregating information from neighboring nodes (Kipf and Welling 2016; Wu et al. 2020). The update of the $l$-th layer in a GNN is expressed by two learnable operations—an *aggregation* function $\text{agg}(\cdot)$ and a *combination* function $\text{update}(\cdot)$. This process can be written as two steps:

First, an aggregation step gathers information from neighboring nodes into a message $\mathbf{m}_{\mathcal{N}(v)}^{(l)}$:

$$\mathbf{m}_{\mathcal{N}(v)}^{(l)} = \text{agg}^{(l)}\left(\left\{\mathbf{h}_u^{(l-1)} \mid u \in \mathcal{N}(v)\right\}\right) \quad (1)$$

Then, an update step combines the message with the node's previous representation to form the new representation:

$$\mathbf{h}_v^{(l)} = \sigma\left(\text{update}^{(l)}\left(\mathbf{h}_v^{(l-1)}, \mathbf{m}_{\mathcal{N}(v)}^{(l)}\right)\right) \quad (2)$$

where $\mathbf{h}_v^{(l)}$ denotes the representation of node $v$ at layer $l$, and $\mathcal{N}(v)$ is the neighbour set of $v$.

### Adversarial Attacks on Graph Neural Networks

Graph adversarial attacks can be summarized as an optimization problem (Zügner, Akbarnejad, and Günnemann 2018). Let $G = (V, E, \mathbf{X})$ be the original graph, and $f_\theta$ be the GNN model. An attacker aims to find a modified graph $G' = (V', E', \mathbf{X}')$ to maximize a certain attack objective function $L_{\text{atk}}$ under the constraint of a perturbation budget $\Delta$. According to the stage at which the attack occurs, it can be divided into poisoning attacks that affect model training (Zügner, Akbarnejad, and Günnemann 2018) and evasion attacks that mislead a fixed model during inference (Feng et al. 2021).

Based on how attackers manipulate graph data, attacks are mainly categorized into two types: graph structure perturbation (Hu et al. 2023; Wu et al. 2024) and node injection attacks (Tao et al. 2021). The former achieves the attack goal by modifying existing edges or node features in the graph, and the perturbed graph has the same node set as the original graph $G$, i.e., $V(G') = V(G)$. The latter realizes the attack by adding new nodes $V_{\text{inj}}$ controlled by the attacker and corresponding edges to the graph. Thus, $G'$ is a supergraph of $G$, satisfying $V(G) \subset V(G')$ and $E(G) \subseteq E(G')$.

In addition, depending on the attacker's knowledge of the target model, attacks can be classified into white-box attacks and black-box attacks. In white-box attacks, the attacker has full knowledge of the model and can utilize its gradients; in the more challenging black-box attacks, the attacker has little knowledge of the model's internal information and usually can only rely on query feedback (Ju et al. 2023).

This paper focuses on one of the most challenging scenarios: evasion node injection attacks in a black-box environment. Our core goal is to maximize the attack effect on the target node set $T \subseteq V$ by injecting malicious nodes and edges into the original graph $G$ to form a modified graph $G'$, under the constraint of a limited number of injected nodes and edges. The objective function of this attack can be formally expressed as maximizing the number of misclassified target nodes:

$$\max_{G'} \sum_{v \in T} I\left(f_{\theta^*}(v, G') \neq y_v\right) \quad (3)$$

where $I(\cdot)$ is an indicator function.

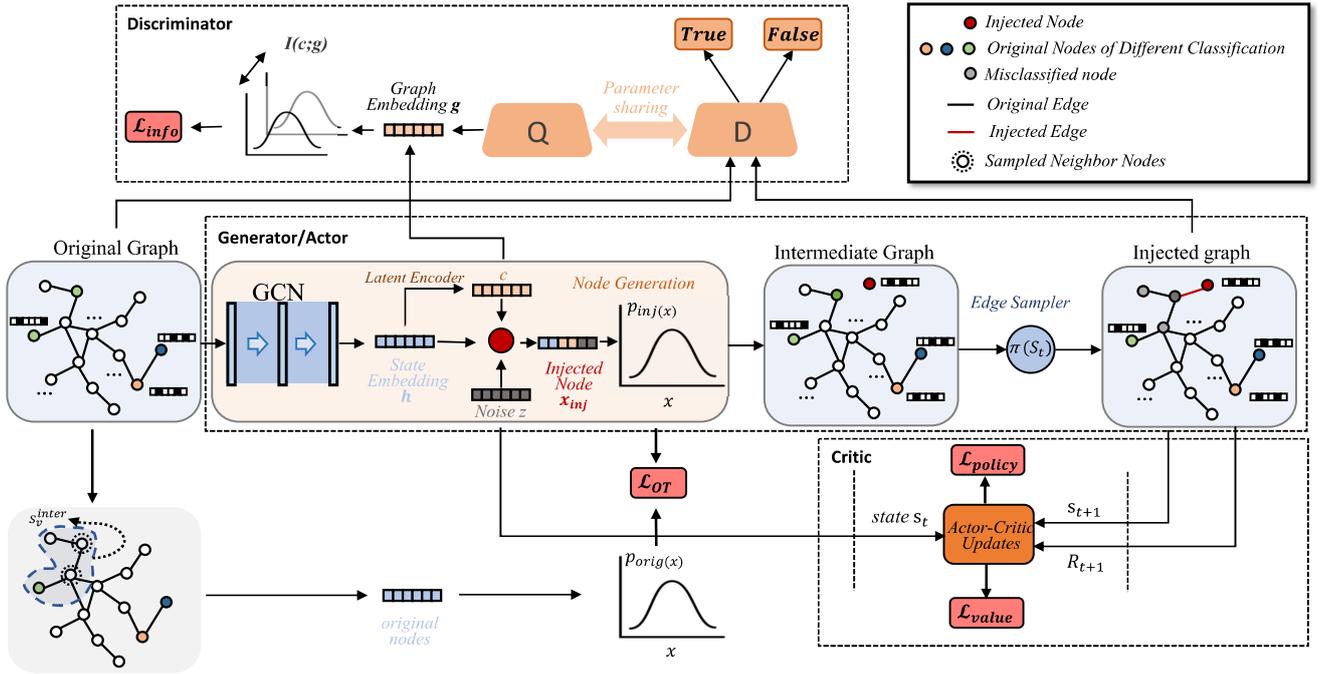

Figure 1: Framework of JANUS

## Methodology

Traditional Graph Injection Attacks are often modeled as static optimization problems. However, this paradigm fails to capture the dynamic and generative nature of the attack process. True stealthiness requires injected nodes and edges to not only evade detection but be statistically and structurally integrated with the original graph's data distribution.

Based on this, we propose a fundamental shift in perspective: treating high-stealth node injection attacks as a Generative Modeling Problem. Our goal is to learn a generator that can generate new nodes along with their incident edges conditioned on the current graph state, such that the modified graph is indistinguishable from the true data distribution of the original graph at both statistical and structural levels.

As shown in Figure 1, the proposed JANUS redefines node injection attacks as a generative modeling task. It is centered on a Generative Adversarial Network (GAN) framework driven by Reinforcement Learning (RL), consisting of a generator $G_\theta$ as an Actor, a discriminator $D_\psi$ for imposing global constraints, and Critic $V_\phi$ for evaluating state values. The entire process is optimized through RL mechanisms to maximize attack effectiveness.

### Local Stealthiness Strategy

To ensure the authenticity of injected nodes at the individual level, we propose a Local Feature Manifold Alignment strategy. Our goal is to ensure that the feature vector $\mathbf{x}_{\text{inj}}$ of the injected node becomes a truly credible sample from the underlying data manifold formed by the features of real nodes in its neighborhood.

We incorporate OT (Peyré and Cuturi 2019). Even when the supports of the empirical distributions of the injected single node and the point cloud of real nodes in its neighborhood do not overlap at all, our local feature manifold alignment still needs to provide smooth and meaningful gradient signals to measure the integration degree of the point-set.

Formally, for each target node, we first establish a fixed reference distribution. We sample a set of features $X_{\text{orig}} = \{\mathbf{x}_{\text{orig}}^1, \ldots, \mathbf{x}_{\text{orig}}^n\}$ from high-degree benign nodes within the target's K-hop neighborhood, as they are more structurally influential. This set forms an empirical probability distribution $P_{\text{orig}} = \frac{1}{n}\sum_{j=1}^{n}\delta_{\mathbf{x}_{\text{orig}}^j}$. Then, at each timestep $t$, the injected feature vector $\mathbf{x}_{\text{inj}}^t$ is regarded as a single-sample Dirac distribution $P_{\text{inj}}^t = \delta_{\mathbf{x}_{\text{inj}}^t}$. The feasible set of the coupling matrix $\gamma$ for transporting the per-timestep distribution $P_{\text{inj}}^t$ to the fixed reference distribution $P_{\text{orig}}$ is:

$$\Gamma(P_{\text{inj}}^t, P_{\text{orig}}) = \left\{ \gamma \in \mathbb{R}_+^{1\times n} \mid \sum_{j=1}^{n}\gamma_j = 1, \gamma_j \geq 0 \right\} \quad (4)$$

where $\gamma_j$ represents the proportion of mass transported from $\mathbf{x}_{\text{inj}}^t$ to a node in $X_{\text{orig}}$. The local stealthiness loss for the single node injected at timestep $t$ is defined as the square of the Wasserstein-2 distance (Peyré and Cuturi 2019) between these two distributions:

$$\mathcal{L}_{\text{OT}}^{X_{\text{inj}}}(\mathbf{x}_{\text{inj}}^t, X_{\text{orig}}) = \min_{\gamma \in \Gamma(P_{\text{inj}}^t, P_{\text{orig}})} \sum_{j=1}^{n} \gamma_j \|\mathbf{x}_{\text{inj}}^t - \mathbf{x}_{\text{orig}}^j\|_2^2 \quad (5)$$

where $\|\cdot\|_2$ denotes the Euclidean norm. In practice, to achieve differentiability, we use the Sinkhorn algorithm (Cuturi 2013) with entropy regularization for solving. The total OT loss term we finally use to guide the generator update is the average of these per-timestep losses over all $N_{\text{inj}}$ injection steps:

$$\mathcal{L}_{\text{OT}} = \frac{1}{N_{\text{inj}}} \sum_{t=1}^{N_{\text{inj}}} \mathcal{L}_{\text{OT}}^{X_{\text{inj}}}(\mathbf{x}_{\text{inj}}^t, X_{\text{orig}}) \qquad (6)$$

### Global Stealthiness Constraints

To achieve macroscopic stealthiness at the graph level, we construct a Global Semantic Structure Alignment strategy. This strategy combines adversarial generation with mutual information-based structural alignment.

**Discriminator** We introduce a discriminator $D_\psi$, whose goal is to distinguish between real graph data and data forged by the generator. We select a powerful Graph Isomorphism Network (GIN) (Xu et al. 2018) as the core architecture. The discriminator's adversarial loss adopts the standard LSGAN (Mao et al. 2017) loss $\mathcal{L}_{\text{adv}}^{(D)}$.

**Sequential Generator** The generator $G_\theta$ of JANUS completes a full node injection action through the sequential operation of two modules:

1. Node Generator: To ensure that the generated features $\mathbf{x}_{\text{inj}}$ conform to the feature conventions of the original graph, the Node Generator first uses K-layer Graph Convolutional Layers (GCLs) to perform message propagation on the K-order subgraph of the target node $v_i$. It then aggregates information via a readout function to construct a context-aware state representation $\mathbf{h}$:

$$\mathbf{h} = \text{Concat}\left(\sum_{v \in \mathcal{S}_i} \mathbf{h}_v^{(K)}, \max_{v \in \mathcal{S}_i} \mathbf{h}_v^{(K)}, \mathbf{h}_{v_i}^{(K)}\right) \qquad (7)$$

where $\mathcal{S}_i$ denotes the set of nodes in the K-hop subgraph of $v_i$, and $\mathbf{h}_v^{(K)}$ is the final K-layer representation for node $v$. Then, the logits $\ell$ for the feature distribution are synthesized by a vector that fuses the normalized $\mathbf{h}$, a latent code $\mathbf{c}$, and additional noise $\mathbf{z}$:

$$\ell = \text{MLP}_{\text{node}}\left(\text{Concat}\left(\text{Normalize}(\mathbf{h}), \mathbf{c}, \mathbf{z}\right)\right) \qquad (8)$$

Finally, the generator produces the feature vector $\mathbf{x}_{\text{inj}}$ by using the Gumbel-Softmax trick(Jang, Gu, and Poole 2016) for discrete features and sampling from a learned Gaussian distribution for continuous ones.

2. Edge Sampler: After generating node features, the Edge Sampler computes a connection probability vector. For each candidate node $v_j$, a connection score $s_j$ is calculated. This score is based on a composite representation of the new feature $\mathbf{x}_{\text{inj}}$ and up-to-date node embeddings ($\mathbf{h}_{v_j}$, $\mathbf{h}_{v_i}$), which are obtained from an internal GNN encoder. A bias term is also added:

$$s_j = \text{MLP}_{\text{edge}}\left(\text{Concat}(\mathbf{h}_{v_j}, \mathbf{h}_{v_i}, \mathbf{x}_{\text{inj}})\right) + \alpha \cdot I_{v_j \in \mathcal{N}(v_i)} \qquad (9)$$

where $I_{v_j \in \mathcal{N}(v_i)}$ is an indicator function and the bias $\alpha$ promotes the formation of structurally more stealthy triangular loops by increasing the connection probability to existing neighbors. Then, these scores are normalized via a softmax function to produce the final probability distribution over all candidate nodes:

$$p(v_j | v_i, \mathbf{x}_{\text{inj}}) = \text{Softmax}_j(s_j) \qquad (10)$$

An edge is finally sampled according to this distribution to complete the connection.

**Latent Coding** To solve the local myopia problem, we introduce structured latent variables $\mathbf{c} = [\mathbf{c}_{\text{disc}}, \mathbf{c}_{\text{cont}}]$ to model discrete semantic categories and continuous attributes, respectively. The sampling distribution of latent variables is dynamically adjusted by a context-aware latent encoder based on the state representation $\mathbf{h}$:

$$\begin{aligned} p_{\text{disc}}, \mu_c, \sigma_c &= \text{LatentEncoder}(\mathbf{h}) \\ \mathbf{c}_{\text{disc}} &\sim \text{Categorical}(p_{\text{disc}}), \quad \mathbf{c}_{\text{cont}} \sim \mathcal{N}(\mu_c, \sigma_c^2) \end{aligned} \qquad (11)$$

To ensure that the generator uses the latent code $\mathbf{c}$ to learn high-order structural semantics of the attack, we maximize the mutual information between $\mathbf{c}$ and a representation of the final generated structure. Since our generator $G$ is a sequential policy conditioned on the state representation $\mathbf{h}$, the structured code $\mathbf{c}$, and unstructured noise $\mathbf{z}$, its final output is a modified subgraph. We obtain a graph-level embedding, denoted as $\mathbf{g}$, by inputting this modified subgraph into the encoder component of our discriminator network $D_\psi$. The mutual information objective is then $I(\mathbf{c}; \mathbf{g})$. Following the variational inference approach from InfoGAN (Chen et al. 2016), the loss is:

$$\mathcal{L}_{\text{info}} = -\mathbb{E}_{\mathbf{h}, \mathbf{c}, \mathbf{z}}\left[\log Q(\mathbf{c} \mid \mathbf{g})\right] \qquad (12)$$

where the auxiliary network $Q$ is trained to recover the latent code $\mathbf{c}$ from a graph-level embedding $\mathbf{g}$. This network $Q$ is integrated into the main discriminator structure $D_\psi$ and shares most of its parameters. For discrete latent codes $\mathbf{c}_{\text{disc}}$, cross-entropy loss is used here. For continuous latent codes $\mathbf{c}_{\text{cont}}$, the negative log-likelihood under the Gaussian distribution predicted by $Q$ is minimized. Finally, the total loss of the discriminator is the sum of the adversarial loss and the information loss:

$$\mathcal{L}_D^{\text{total}} = \mathcal{L}_{\text{adv}}^{(D)} + \mathcal{L}_{\text{info}} \qquad (13)$$

### RL-Driven Attack Optimization

To address challenges such as the non-differentiability of black-box objectives and the sequential nature of the decision-making process, we model this problem as a Markov Decision Process (MDP). We adopt an Actor-Critic architecture (Konda and Tsitsiklis 1999) to solve this MDP. The architecture consists of two parts: the Actor network, a policy network $\pi_\theta$, which is also our generator, responsible for outputting an action $\mathbf{a}_t$ based on the current state $\mathbf{s}_t$; and the Critic network, a value network $V_\phi$ that estimates the state value $V_\phi(\mathbf{s}_t)$ and supplies low-variance guidance signals for the actor's policy updates. The reward is the increase in the victim's classification loss, supplemented by a discrete reward for successful misclassification.

We use a policy loss $\mathcal{L}_{\text{policy}}$ for the Actor loss, aiming to maximize the attack success rate, which is defined as:

$$\mathcal{L}_{\text{policy}} = \sum -\log \pi_\theta(\mathbf{a}_t \mid \mathbf{s}_t) A_t \qquad (14)$$

where $A_t = R_t - V_\phi(\mathbf{s}_t)$ is the advantage function. We use Smooth L1 loss to calculate the value loss for optimizing the Critic, in the form of:

$$\mathcal{L}_{\text{value}} = \sum \text{SmoothL1Loss}(V_\phi(\mathbf{s}_t), R_t) \qquad (15)$$

**Unified Training Objective** The training of JANUS involves a multi-objective optimization for the generator (Actor). The generator's primary goal is to maximize the attack reward while satisfying the dual-stealthiness constraints. Its unified loss function is:

$$\mathcal{L}_G^{\text{total}} = \mathcal{L}_{\text{policy}} + \mathcal{L}_{\text{adv}}^{(G)} + \mathcal{L}_{\text{info}} + \lambda_{\text{OT}} \mathcal{L}_{\text{OT}} \qquad (16)$$

where the hyperparameter $\lambda_{\text{OT}}$ is used to balance attack effectiveness and local stealthiness. The training alternates between updating the generator, the discriminator, and the critic, as detailed in Algorithm 1.

---

**Algorithm 1: Training Algorithm of JANUS**

1: **Input:** Graph $G$, victim model $M$, training rounds $N$, attack budget $\Delta$
2: **Output:** Optimal attack policy $\pi_\theta^*$
3: Initialize Generator (Actor) $\theta$, Discriminator $\psi$, and Critic $\phi$.
4: **for** epoch = 1 to $N$ **do**
5:   Get initial state $\mathbf{s}_0$ from the environment.
6:   **for** $t = 0$ to $\Delta - 1$ **do**
7:     Sample action $\mathbf{a}_t = (\mathbf{x}_{\text{inj}}, \mathbf{e}_{\text{inj}})$ from policy $\pi_\theta(\cdot \mid \mathbf{s}_t)$.
8:     Calculate local stealthiness cost $\mathcal{L}_{\text{OT}}$ for $\mathbf{x}_{\text{inj}}$.
9:     Execute action $\mathbf{a}_t$, get next state $\mathbf{s}_{t+1}$ and reward $r_t$.
10:    Store transition $(\mathbf{s}_t, \mathbf{a}_t, r_t, \mathbf{s}_{t+1})$ in buffer $B$.
11:   **end for**
12:   Update Critic $\phi$ by minimizing $\mathcal{L}_{\text{value}}$ on a batch from $B$.
13:   Update Generator (Actor) $\theta$ by minimizing $\mathcal{L}_G^{\text{total}}$ on a batch from $B$.
14:   Update Discriminator $\psi$ by minimizing $\mathcal{L}_D^{\text{total}}$ on a batch from $B$.
15: **end for**

---

## Experiments

In this section, our experimental design aims to answer the following core research questions: (RQ1) How does the attack success rate of JANUS compare to existing baselines? (RQ2) Can JANUS maintain strong attack capability when facing mainstream graph defense mechanisms? (RQ3) What is the stealthiness performance of JANUS's injected nodes from both quantitative and visual perspectives? (RQ4) Do the core components in JANUS make essential contributions to its final performance?

### Experimental Setup

**Datasets.** We conducted experiments on eight well-recognized benchmark datasets, including citation networks Cora, Citeseer, Pubmed (Sen et al. 2008), and Cora-ML (Bojchevski and Günnemann 2017); the co-purchase network Amazon Photo (Shchur et al. 2018); the Bayesian network UAI (Wang et al. 2018); and the Open Graph Benchmark OGB-Products (Hu et al. 2020). These datasets cover multiple domains and feature spaces, which are sufficient to fully verify our method's performance. For UAI and Cora-ML, we adopted a random split of 20%/20%/60% for training, validation, and testing. For other datasets, we used the same subgraphs and data splits as $G^2A2C$ (Ju et al. 2023) to ensure a fair comparison. Detailed statistics are shown in Table 1.

| Dataset | Node | Edge | Class | Dim. |
|---|---|---|---|---|
| Datasets with Discrete Feature Space | | | | |
| Cora | 2,708 | 5,429 | 7 | 1,433 |
| Citeseer | 3,327 | 4,732 | 6 | 3,703 |
| Am. Photo | 7,650 | 119,043 | 8 | 745 |
| Uai | 3,067 | 28,311 | 19 | 4,973 |
| Datasets with Continuous Feature Space | | | | |
| Pubmed | 19,717 | 44,338 | 3 | 500 |
| Wiki. CS | 11,701 | 216,123 | 10 | 300 |
| OGB-Prod. | 10,494 | 77,656 | 35 | 100 |
| Cora_ml | 2,995 | 4,208 | 7 | 2,879 |

Table 1: Statistical Information of Experimental Datasets.

**Baselines.** To comprehensively evaluate the performance of JANUS, we compare it with a series of state-of-the-art node injection attack methods. These baselines include reinforcement learning-based black-box methods, such as NIPA (Sun et al. 2020) and $G^2A2C$ (Ju et al. 2023), as well as proxy gradient-based methods like G-NIA (Tao et al. 2021), TDGIA (Zou et al. 2021), and AFGSM (Wang et al. 2020). For methods that require a white-box setting, we uniformly trained a two-layer GCN as their proxy model. In addition, we introduced camouflage enhancement frameworks HAO (Chen et al. 2022) and CANA (Tao et al. 2023) combined with TDGIA as stronger stealthy attack baselines.

**Implementation Details.** For all baselines, we adopt their default parameter settings or use implementations from the DeepRobust library (Li et al. 2020). For our proposed JANUS framework, we use the Adam optimizer with learning rates of 1e-4 for the generator/critic and 1e-5 for the discriminator. The GCN hidden dimension is 128. The hyperparameter $\lambda_{\text{OT}}$ is selected via grid search in the range [0.1, 1] with a step of 0.1. The dimension of discrete latent codes is the same as the number of dataset categories, and the dimension of continuous latent codes is set to 15. Training is conducted for a maximum of 10,000 epochs with an early stopping mechanism with a patience of 15. All experiments are conducted on a server equipped with an NVIDIA A800 GPU and an Intel Xeon Gold 6348 CPU; each experiment is repeated ten times and the average results are reported.

| Attacker | Discrete Feature Space | | | | Continuous Feature Space | | | |
|---|---|---|---|---|---|---|---|---|
| | Cora | Citeseer | Uai | Am. Photo | OGB-Prod. | Cora_ml | Pubmed | Wiki CS |
| Clean | 19.1 | 25.1 | 27.7 | 17.8 | 23.2 | 14.4 | 21.9 | 18.3 |
| NIPA | $19.7_{\pm 0.2}$ | $25.2_{\pm 0.1}$ | $27.9_{\pm 0.3}$ | $17.8_{\pm 0.}$ | $24.1_{\pm 0.3}$ | $15.1_{\pm 0.2}$ | $21.9_{\pm 0.}$ | $19.2_{\pm 0.4}$ |
| AFGSM | $26.1_{\pm 3.2}$ | $39.9_{\pm 3.5}$ | $30.6_{\pm 2.1}$ | $32.3_{\pm 1.9}$ | $75.2_{\pm 0.8}$ | $63.5_{\pm 1.1}$ | $66.0_{\pm 1.3}$ | $75.2_{\pm 0.9}$ |
| G-NIA | $24.5_{\pm 2.8}$ | $40.5_{\pm 3.0}$ | $32.3_{\pm 2.5}$ | $25.0_{\pm 1.8}$ | $97.1_{\pm 0.5}$ | $69.3_{\pm 1.1}$ | $68.1_{\pm 1.0}$ | $79.3_{\pm 1.5}$ |
| TDGIA | $31.2_{\pm 2.5}$ | $44.2_{\pm 2.8}$ | $32.5_{\pm 2.4}$ | $\underline{34.1_{\pm 1.3}}$ | $95.3_{\pm 0.4}$ | $68.7_{\pm 1.2}$ | $70.9_{\pm 0.7}$ | $84.2_{\pm 1.1}$ |
| G$^2$A2C | $39.1_{\pm 2.9}$ | $50.3_{\pm 3.2}$ | $34.3_{\pm 2.2}$ | $33.3_{\pm 1.5}$ | $96.0_{\pm 0.4}$ | $72.6_{\pm 1.3}$ | $73.4_{\pm 0.9}$ | $86.1_{\pm 0.9}$ |
| JANUS (ours) | **$60.7_{\pm 3.1}$** | **$66.9_{\pm 3.1}$** | **$46.4_{\pm 1.6}$** | **$41.0_{\pm 1.1}$** | **$98.6_{\pm 0.2}$** | **$79.3_{\pm 0.9}$** | **$89.5_{\pm 1.2}$** | **$91.8_{\pm 1.2}$** |
| Avg. ↑ | 21.6 | 16.6 | 12.1 | 6.9 | 1.5 | 6.7 | 16.1 | 5.7 |

Table 2: Misclassification rates (%) on two-layer GCN under single-node and single-edge injection attacks. The best results are in **bold** and the second-best are underlined. We report the mean and standard deviation over 10 runs with different seeds. The improvements of our method over all baselines are verified to be statistically significant via paired t-tests ($p < 0.01$).

### Analysis of Attack Effectiveness

To evaluate the attack effectiveness of JANUS, we first conducted attacks on a two-layer GCN victim model under the highly challenging single-node and single-edge injection setting. As reported in Table 2, JANUS achieves the highest misclassification rate across all datasets, significantly outperforming all baseline methods. These improvements are statistically significant across all datasets($p < 0.01$). On the widely used Citeseer and Cora_ml benchmarks, for instance, JANUS achieves misclassification rates of 66.9% and 79.3% respectively, far exceeding all baseline methods. This superior performance, even under the strictest budget, demonstrates a key insight of our work: stealthiness is not a trade-off against efficacy but a direct enabler of it. The dual-stealthiness mechanism compels JANUS to discover fundamentally more deceptive and effective attack vectors within the natural data manifold, thus comprehensively answering RQ1.

### Robustness Against Defenses

To evaluate the robustness of JANUS in realistic adversarial environments, we test its effectiveness against two mainstream defense models: GNNGuard (Zhang and Zitnik 2020) and FLAG (Kong et al. 2020). In a large-scale attack on Citeseer, we inject 1000 malicious nodes, each with 2 edges, while on OGB-Products, we inject 2099 nodes, each with 7 edges. As shown in Table 3, JANUS demonstrates superior robustness, consistently outperforming baselines. This advantage stems from producing attack patterns fundamentally harder for defenses to neutralize. By ensuring both local feature authenticity and global structural consistency, its natural features evade feature-level scrutiny, while its globally coherent structure bypasses topological defense mechanisms like GNNGuard's attention, thus providing a clear answer to RQ2.

### Stealthiness Evaluation

We first evaluate stealthiness using quantitative metrics to answer RQ3. We adopt two common metrics where lower is better: Closest Attribute Distance (CAD) (Tao et al. 2023), which measures feature similarity to the nearest original node, and Smoothness (Dong, Zhang, and Wang 2023), which measures feature consistency with connected neighbors. The results, presented in Figure 2, show that JANUS consistently achieves the best performance on the OGB-Products dataset. It outperforms all baselines across both metrics, establishing its state-of-the-art (SOTA) stealthiness in terms of feature and structural similarity.

| Attacker | Backbone | Citeseer | OGB-Prod. |
|---|---|---|---|
| TDGIA | FLAG | 43.7 | 75.7 |
| | GNNGuard | 45.8 | 80.5 |
| TDGIA +CANA | FLAG | 44.0 | 82.2 |
| | GNNGuard | 48.9 | 84.1 |
| TDGIA +HAO | FLAG | 47.9 | 87.3 |
| | GNNGuard | 49.6 | 90.9 |
| JANUS | FLAG | **50.5** | **98.1** |
| | GNNGuard | **60.4** | **93.2** |

Table 3: Misclassification Rates (%) under Defense Models.

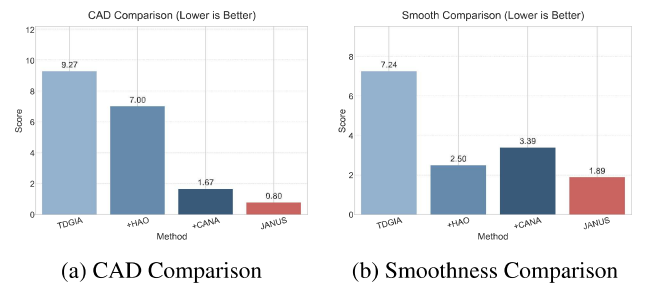

(a) CAD Comparison      (b) Smoothness Comparison

Figure 2: Quantitative stealth metrics (↓ lower is better) on the OGB-Products dataset.

To provide more intuitive evidence, we present a t-SNE visualization(Maaten and Hinton 2008) in Figure 3 that offers two complementary insights. First, the IForest detec-tion AUCs reported in the figure's captions serve to validate our local feature manifold alignment. The baseline's high AUC of 0.90 indicates its features are detectable, whereas the near-random 0.48 AUC for JANUS proves its superior feature-level stealth. Second, the visual distribution of the nodes validates our global semantic structure alignment. As shown in the visualization, the baseline's injected nodes form distinct clusters. In stark contrast, JANUS's nodes are uniformly distributed and seamlessly integrated with the original node distribution. Crucially, JANUS achieves this comprehensive, SOTA stealth while maintaining its superior attack success rate (Table 2), thus establishing a more advanced Pareto front.

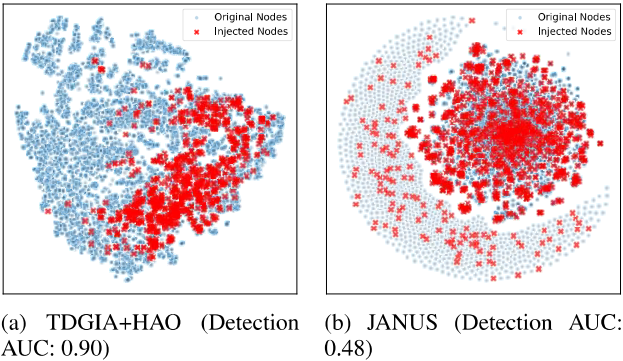

(a) TDGIA+HAO (Detection AUC: 0.90)   (b) JANUS (Detection AUC: 0.48)

Figure 3: t-SNE visualization of injected nodes (red) versus original nodes (blue) on the OGB-Products dataset.

## Ablation Study

To verify the effectiveness of each core component in JANUS (RQ4), we designed three ablation variants: **JANUS w/o local**, which removes the local OT alignment for node features; **JANUS w/o global**, which removes the global stealthiness module composed of adversarial learning and latent coding; and a pure **RL-only** baseline that only uses reinforcement learning for node and edge generation. The results in Table 4 demonstrate the contribution of each component. Removing the local feature manifold alignment (w/o local) results in a significant performance drop, underscoring the criticality of generating natural features to bypass feature-based detection, a contribution especially visible against the robust GNNGuard defense.

| Variant | GCN | GNNGuard | FLAG |
|---|---|---|---|
| JANUS | **66.9** | **56.7** | **46.8** |
| JANUS w/o local | 58.3 | 49.7 | 43.2 |
| JANUS w/o global | 52.6 | 47.9 | 40.7 |
| RL-only | 43.2 | 39.2 | 36.3 |

Table 4: Attack success rate (%) of JANUS variants on Citeseer.

Notably, removing the global semantic structure alignment (w/o global) leads to an even more severe performance degradation. This highlights its crucial role in preventing the accumulation of locally plausible injections into a structurally anomalous pattern at the macroscopic level. By forcing the injected structures to adhere to the graph's underlying semantics, it overcomes the local myopia problem, which is particularly vital for evading advanced defenses adept at identifying structural inconsistencies. The pure reinforcement learning benchmark (RL-only) performed the worst, confirming that the attack's success is driven primarily by the stealthiness constraints. The superior performance of JANUS confirms both constraints are indispensable and synergistic, thus answering RQ4.

## Related Work

Adversarial attacks on GNNs are primarily categorized into two paradigms: Graph Manipulation Attacks (GMA) and Graph Injection Attacks (GIA). GMA methods (Zügner, Akbarnejad, and Günnemann 2018) perturb existing edges or features but are often impractical due to the requirement of direct data modification. In contrast, GIAs (Sun et al. 2020), which only add new nodes and connections, represent a more feasible threat model.

However, existing GIA methods exhibit significant limitations. A prominent line of work, including G-NIA (Tao et al. 2021) and TDGIA (Zou et al. 2021), relies on gradients from surrogate models. The effectiveness of these methods degrades significantly when the surrogate model diverges from the true victim architecture. Furthermore, their focus on local feature simulation without ensuring global consistency often leads to a local myopia problem. More recently, $G^2A2C$ (Ju et al. 2023) introduced a reinforcement learning (RL) approach to operate in a gradient-free, black-box setting. Nevertheless, it still relies on heuristic constraints for stealthiness and does not explicitly enforce the naturalness of the global graph semantics. To address these deficiencies, this paper proposes JANUS, which integrates optimal transport for local feature manifold alignment and mutual information maximization for global structural semantics into a unified RL framework to simultaneously enhance attack effectiveness and stealthiness.

## Conclusion

In this work, we study the challenging problem of stealthy black-box node injection attacks against GNNs. To address this problem, we propose JANUS, a generative attack framework centered on a novel dual-stealthiness constraint mechanism. We reframe the attack as a generative modeling problem and model it as a sequential decision process, optimized by a reinforcement learning agent. This agent is guided by two key principles: local feature manifold alignment via optimal transport and global structural consistency via mutual information maximization, ensuring both feature and structural naturalness. Through comprehensive experiments, we demonstrate that JANUS significantly outperforms existing SOTA methods in both attack success rate and multiple stealthiness metrics. Furthermore, it maintains high effectiveness even when facing advanced defense mechanisms, proving the superiority of its holistic approach to stealth.